\documentclass{article}

\usepackage[preprint]{neurips_2023}

\usepackage[utf8]{inputenc} 
\usepackage[T1]{fontenc}    
\usepackage{hyperref}       
\usepackage{url}            
\usepackage{booktabs}       
\usepackage{amsfonts}       
\usepackage{nicefrac}       
\usepackage{microtype}      
\usepackage{xcolor}         

\usepackage{graphicx} 
\usepackage{svg}
\usepackage{wrapfig}
\usepackage{lmodern}
\usepackage{makecell}

\title{NYONIC TECHNICAL REPORT}

\author{%
  Junfeng Tian, Rui Wang, Cong Li, Yudong Zhou, Jun Liu, Jun Wang \\
  \texttt{nyonic.ai} \\
}

\begin{document}

\maketitle

\begin{abstract}
  This report details the development and key achievements of our latest language model designed for custom large language models. The advancements introduced include a novel Online Data Scheduler that supports flexible training data adjustments and curriculum learning. The model's architecture is fortified with state-of-the-art techniques such as Rotary Positional Embeddings, QK-LayerNorm, and a specially crafted multilingual tokenizer to enhance stability and performance.  Moreover, our robust training framework incorporates advanced monitoring and rapid recovery features to ensure optimal efficiency. Our Wonton 7B model has demonstrated competitive performance on a range of multilingual and English benchmarks. Future developments will prioritize narrowing the performance gap with more extensively trained models, thereby enhancing the model's real-world efficacy and adaptability.
  
  GitHub: \url{https://github.com/nyonicai/nyonic-public}
\end{abstract}

\section{Introduction}


This report details the development and release of our pre-trained 7B base model checkpoint and one fine-tuned model for chat. As we plan to continuously train and improve our model, we will release more base model checkpoints in the future, as well as fine-tuned models for different use cases.

Our training infrastructure is mainly built on open source PyTorch framework, incorporating various newly developed features, e.g., flash attention(\citep{flashattn}), RoPE(\cite{rope}), QK-LayerNorm(\citep{dehghani2023scaling}), etc.

The main contributions of this work can be summarized as follows:
\begin{itemize}
  \item We build an online data scheduler and multiplexer for flexible training data mixing.
  \item We train our own multilingual tokenizer to incorporate several widely-spoken languages.
  \item We monitor various intermediate metrics during the training and apply several normalization and regularization techniques to increase the stability.
  \item We consolidate our infrastructure to speed up the resuming after interruptions.
  \item We fine tuned the pre-trained model with open SFT datasets, as well as our private datasets.
  \item We enabled a few inference scenarios and provided deployment methods.
\end{itemize}

This report aims to provide a comprehensive overview of LLM training, development, SFT, inference, and deployment technology, which can benefit the community in the creation of more LLMs and the development of a wide range of real-world applications.

\section{Training}

\subsection{Data}

The efficacy and performance of language models are significantly influenced by the quality and diversity of the data used in training. An effective pretraining dataset must encompass a broad spectrum of types, domains, and tasks to ensure comprehensive linguistic coverage. 

\begin{figure}[t]
\centering
\includegraphics[width=0.95\textwidth]{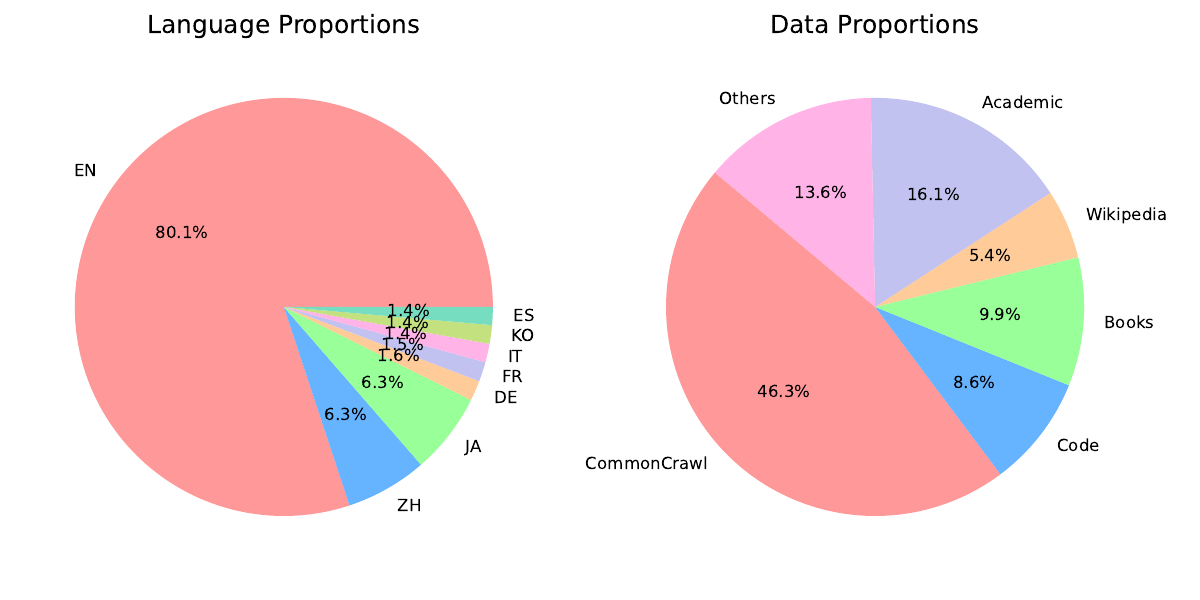} 
\caption{Language and data proportions in the training set.} 
\label{fig:data}
\end{figure}

Our curated dataset is specifically designed to meet these criteria, incorporating sources such as the Common Crawl, books, Wikipedia, code, and academic papers. Moreover, our dataset is multilingual, encompassing English, Chinese, German, French, Italian, Spanish, Japanese, and Korean, and with a significant portion of the data being in English. Figure  \ref{fig:data} illustrates the mixture of data and the percentage they represent in the training set.



\subsection{Data Engineering}

\begin{figure}[t]
\centering
\includegraphics[width=0.95\textwidth]{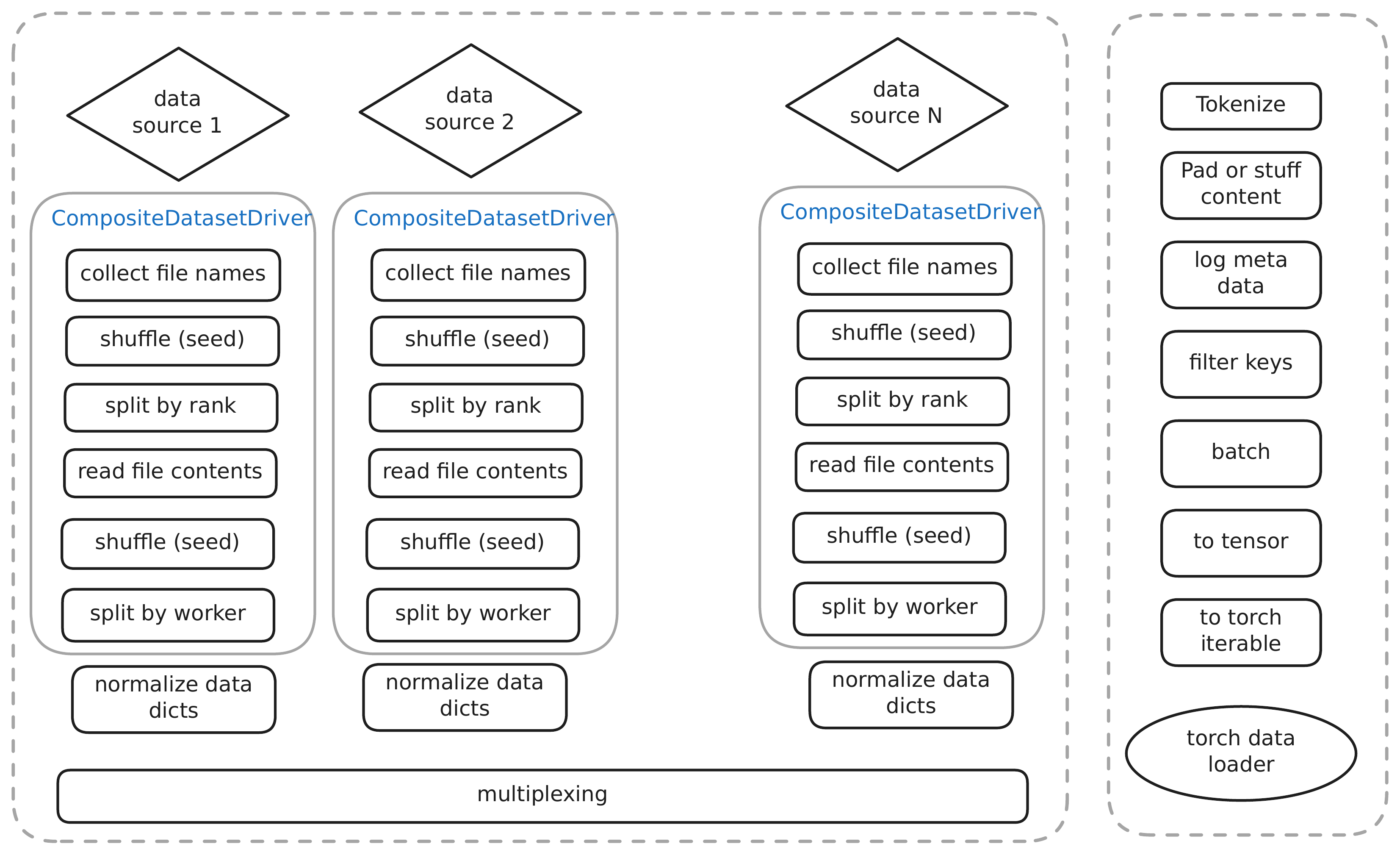} 
\caption{The data flow architecture of the Online Data Scheduler.} 
\label{fig:example}
\end{figure}

In providing customized large model solutions for specific industries, the traditional approach involves offline conversion of data into token indices to create corresponding data indices. However, recent studies (\citep{xie2023doremi}) have shown that compared to uniformly sampling data sources, carefully selecting training data and undergoing continual pre-training can significantly enhance model accuracy, further validating the effectiveness of data scheduling strategies.

Nevertheless, these  methods still rely on offline data processing and lack mechanisms for online adjustments and real-time feedback during the training process. Restarting training not only consumes substantial computational resources but also does not necessarily yield better training outcomes.

To address this issue, we propose an innovative \textbf{Online Data Scheduler}, which integrates an online data stream with a data scheduler. The Online Data Scheduler offers several benefits:
\begin{itemize}
    \item It enables flexible implementation of data mixing since it eliminates the need for offline conversion of data into token indices. By utilizing multiple workers in the data loader, it can concurrently process data and perform model training, without reducing training efficiency.
    \item  The system allows for training flexibility across different stages through curriculum learning. The model is able to avoid wasting time on simple data that it has already mastered, focusing more attention on data that is challenging to learn. This targeted training approach can accelerate model convergence, reduce unnecessary computations, and thereby save time and resources.
    \item Real-time feedback is possible. The scheduler can dynamically adjust data ratios based on the model's real-time training loss, allowing for immediate adjustments based on feedback.
    \item  The model is capable of online learning, enabling it to adapt to new data. This allows the model to learn within a continuously changing data stream, aligning more closely with the ongoing learning processes observed in the real world.
\end{itemize}


The data flow for training large models is efficiently divided into three stages. 
\begin{enumerate}
    \item \textbf{Data Preparation} involves collecting, shuffling, and distributing file names across different ranks, with each rank responsible for reading file contents. 
    \item In the \textbf{Data Processing} stage, data is further shuffled, split among workers, normalized, and tokenized, ensuring consistency and readiness for model input.
    \item  The final stage, \textbf{Batch Preparation}, involves padding the data to uniform lengths, logging metadata, and organizing it into batches that are converted to tensors and torch iterables for training.
\end{enumerate}
This streamlined process ensures that data is methodically prepared and optimized for effective model training. In particular, we present the key components of our online data scheduler, which facilitate the integration of data from various sources in a configurable manner.
\begin{itemize}
    \item \textbf{Multiplexing}: To facilitate data mixing from various sources in a configurable manner, a Multiplexer is integrated into the real data loading pipeline. This allows for the combination or multiplexing of processed data streams from different sources into a single stream, with configurable mixture ratios.
    \item \textbf{Content Stuffing}: Following \citep{gpt3}, during training, we consistently use sequences that fill the entire 2048 token context window. When individual documents are shorter than 2048 tokens, multiple documents are packed into a single sequence to enhance computational efficiency. These sequences are further augmented with additional length information, which enables them to be processed in the CUDA kernel without the need for sequence-specific masking.
    \item \textbf{Data Resuming}: Given  that the list of files assigned to each rank remains constant for each dataset, we choose to save the processing status of files on each rank. This approach enables us to bypass files that have already been processed when resuming training. Currently, our focus is on maintaining lightweight control over the resumption process at the file level, rather than managing it with finer granularity at the record level.
\end{itemize}

\subsection{Tokenization}



\begin{figure}[t]
\centering
\includegraphics[width=\textwidth]{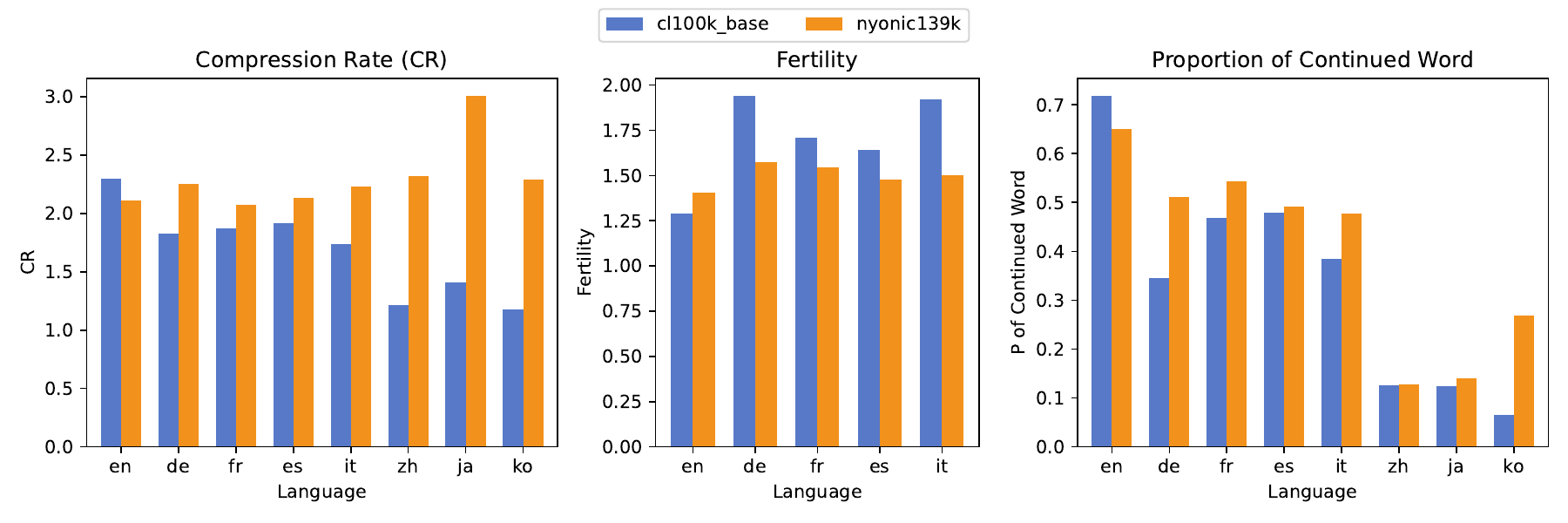} 
\caption{Comparative analysis of multilingual tokenization metrics across different languages.} 
\label{fig:tokenization_CR}
\end{figure}

The design of vocabulary plays a significant role in the training efficiency and downstream task performance of multilingual large language models. We employ the byte-pair encoding (BPE) algorithm, as described by \citep{bpe}, using the SentencePiece implementation (\citep{kudo-richardson-2018-sentencepiece}). To optimize performance with datasets containing abundant whitespace, such as code, we assign one special token for each sequence of whitespace ranging from 1 to 24 characters. Additionally, in line with \citep{llama, llama2}, we have segmented numbers into individual digits.

Selecting an optimal vocabulary size is of paramount importance for tokenizers, as a size that is too small may impede transformer computations, while a size that is too large may hinder the model's capacity to derive meaning from sequences. We have identified the most commonly used languages by our customers, including English, Simplified Chinese, Traditional Chinese, German, French, Italian, Spanish, Japanese, and Korean. Our data sources encompass Common Crawl, Wiki, and various coding repositories. The final vocabulary size of the tokenizer, as determined by extensive experimentation, is approximately 139,000. 

In Figure  \ref{fig:tokenization_CR}, we present a comprehensive evaluation of the multilingual tokenizer. This detailed comparison examines several critical metrics that are essential for understanding tokenizer efficiency. These metrics include the compression rate, which measures the compactness of tokenized data; fertility, which assesses the breakdown of words into subwords; and the proportion of continued words, which evaluates the segmentation consistency across various languages. The results of our analysis demonstrate that our tokenizer achieves superior compression efficiency in most languages. This indicates that it has the potential to significantly reduce operational costs by using fewer tokens to transmit equivalent or greater amounts of information.

\subsection{Architecture Highlights}

Our model, Wonton 7B, is built on a transformer architecture (\citep{transformer}), with its main parameters summarized in Table \ref{tab:param}. Compared to GPT3 (\citep{gpt3}), Wonton 7B incorporates a number of enhancements, which are outlined below.

\paragraph{Rotary Positional Embeddings (RoPE)} (\citep{rope}): Unlike traditional absolute positional embeddings, we implement rotary positional embeddings in each layer. RoPE has gained widespread acceptance and has shown effectiveness in modern large language models (\citep{llama2, jiang2023mistral, bai2023qwen}).

\paragraph{Pre-Norm \& QK-LayerNorm}: We employ pre-normalization, a prevalent strategy that enhances training stability over post-normalization (\citep{xiong2020layer}). Traditional layer normalization (\citep{layer_norm}) is used, and it is also applied to the queries and keys before the dot-product attention process (\citep{dehghani2023scaling}) to prevent excessively large values in attention logits.

\paragraph{Max-z Loss}: Building on the auxiliary z-loss from PaLM (\citep{palm}) and the max-z loss concept (\citep{baichuan2}), we employ these techniques to regulate logits values. This ensures more stable training and robust inference across various hyper-parameters. Max-z loss, in particular, offers greater effectiveness with a reduced memory demand, making it our default choice for model training.

\begin{wraptable}{r}{5cm}
{\ttfamily
\centering
\begin{tabular}{cc}
\toprule
\textbf{Parameter} & \textbf{Value} \\
\midrule
params & 6.7B \\
dimension & 4,096 \\
$n$ heads & 32 \\
$n$ layers & 32 \\
context len & 2,048 \\
vocab size & 139,776 \\
learning rate & $3.0e^{-4}$ \\
batch size & 1M \\
\bottomrule
\end{tabular}
}
\caption{Model architecture.}\label{tab:param}
\end{wraptable}

\subsection{Training Hyperparameters\label{trainparameters}}


The models are trained using the AdamW optimizer (\citep{adamw}), with the hyper-parameters set at $\beta_1 = 0.9$ and $\beta_2 = 0.95$. We implement a weight decay of $0.1$ and apply gradient clipping at $1.0$. Following the approach described in \citep{gpt3}, we utilize a cosine decay schedule with warmup. The warmup period consists of $2,000$ steps, followed by a decay to 10\% of the maximum learning rate.

Following the approach described in \citep{radford2019language}, we initialize the linear and embedding weights using a normal distribution and scale the up-projection layers at initialization by a factor of $1/\sqrt{N}$ where N represents the number of layers. This scaling is designed to optimize the variance of the outputs across different layers, facilitating more stable training.

\subsection{Training Infrastructures}

For training the Nyonic-7B model, we utilize PyTorch (\citep{PyTorch}) and DeepSpeed ZeRO2 (\citep{deepspeed}), distributing the optimizers states across 128 NVIDIA A800 GPUs interconnected via InfiniBand. To enhance training efficiency, we incorporate FlashAttention (\citep{flashattn}) and xformers (\citep{xFormers2022}), achieving a training speed of approximately 3,000 tokens per GPU per second. All models are trained using BFloat16 mixed precision to ensure training stability. 

We introduce additional metrics to monitor the model training process. By integrating these metrics, we aim to gain deeper insights into our model's internal dynamics and improve its performance and stability across a range of tasks.
\begin{itemize}
    \item \textbf{Max Attention Logits}: This measures the maximum value of the attention logits within each attention block, providing insight into the attention distribution's extremities.
    \item \textbf{Mean Query Norm}: By averaging the norms of the query vectors, this metric assesses the average strength or magnitude of the queries before they interact with the keys, which can indicate the overall query signal strength across different layers.
    \item \textbf{Output Logit Mean (Pre-Softmax)}: This metric calculates the mean of the output logits just before the softmax activation, which can help in understanding the pre-activation distribution of the logits.
    \item \textbf{Root Mean Square of Gradient of the First Layer of MLP}: Monitoring the root mean square of the gradient for the first layer of the multi-layer perceptron (MLP) within each block can provide early indications of potential issues with gradient vanishing or exploding, which are critical for maintaining training stability.
    \item \textbf{Block Output RMS}: The root mean square (RMS) of each block’s output offers a measure of the output signal's consistency and variability, which can be crucial for diagnosing model behavior during both training and inference phases.
\end{itemize}

Our training framework supports minute-level training resumption thanks to our implemented online data scheduler. Additionally, it accommodates training resumption from various distributing configurations to adapt to dynamic changes in the computing cluster.
    
\subsection{Experiment Results}

\paragraph{Basic Capabilities}

We select the most common evaluation sets in each category as follows:
\begin{itemize}
    \item \textbf{Lambada} (\citep{lambada}): The LAMBADA benchmark features around 10,000 passages from the BooksCorpus, where participants are tasked with predicting a missing word in the contextually rich final sentence of each passage.
    \item \textbf{WinoGrande} (\citep{winogrande}): A large-scale dataset designed to challenge models with complex commonsense reasoning through refined pronoun disambiguation tasks.
    \item \textbf{HellaSwag} (\cite{hellaswag}): This dataset pushes the limits of models' commonsense reasoning capabilities by requiring them to complete scenarios in a contextually appropriate way.
    \item \textbf{PIQA} (\citep{piqa}): Introduces a benchmark for physical commonsense reasoning, testing models’ abilities to understand and predict physical interactions in diverse settings.
    \item \textbf{RACE} (\citep{race}): A robust dataset providing a range of reading comprehension questions based on academic texts from middle and high school levels.
    \item \textbf{BoolQ} (\citep{boolq}): Focuses on yes/no question answering, with questions naturally derived from Google searches, testing models’ abilities to interpret and evaluate straightforward queries against provided texts.
\end{itemize}


\begin{table}[t]
\centering
\setlength{\tabcolsep}{8pt} 
\begin{tabular}{lcccccc}
\toprule
Task & \makecell[c]{Lambada \\(5153)} & \makecell[c]{WinoGrande \\ (1267)} & \makecell[c]{HellaSwag \\ (10042)} & \makecell[c]{PIQA \\ (1838)} & \makecell[c]{RACE \\ (1045)} & \makecell[c]{BoolQ \\ (3270)}\\
Metric & ppl $\downarrow$ & acc & acc & acc & acc & acc \\
\midrule
Mistral 7B & 3.7771 & 0.7364 & 0.6124 & 0.8074 & 0.4096 & 0.8376 \\
Pythia 7B & 6.9210 & 0.6109 & 0.4811 & 0.7514 & 0.3684 & 0.6358 \\
Wonton 7B & 6.2973 & 0.6456 & 0.4777 & 0.7503 & 0.3684 & 0.6685 \\
\bottomrule
\end{tabular}
\caption{Performance of Wonton 7B and other open source models on the benchmarks. All models were re-evaluated on all metrics with our evaluation pipeline for accurate comparison.}\label{tab:result_basic}
\end{table}

Table \ref{tab:result_basic} lists the performance of Wonton 7B and the other open source models on the benchmarks. Wonton 7B consistently outperforms Pythia 7B across various metrics, indicating more efficient use of its training data and better optimization strategies. However, when compared to Mistral 7B, which was trained on a significantly larger corpus, Wonton 7B still lags behind, especially in tasks that require a deep understanding of language nuances and complex reasoning. These results underscore the effectiveness of Wonton 7B against comparable models like Pythia but also highlight the challenges it faces in reaching the performance levels of more extensively trained models such as Mistral 7B. This comparison not only reflects Wonton 7B's strengths but also points towards potential areas for future improvement to bridge the gap with top-tier models.

\paragraph{Multilingual Understanding}

To assess the capabilities of Wonton 7B, we utilized various benchmark datasets that challenge models to demonstrate comprehension and reasoning across multiple languages. The datasets employed include:
\begin{itemize}
    \item \textbf{Belebele} (\citep{belebele}): Tests models on tasks across eight different languages—English, German, French, Italian, Spanish, Chinese, Japanese, and Korean.
    \item \textbf{XNLI} (\citep{xnli}): Focuses on natural language inference in three languages, assessing how well models understand and interpret text across language barriers.
    \item \textbf{XStoryCloze} (\citep{xstorycloze}): Measures models' ability to logically complete stories in English, Spanish, and Chinese, testing narrative understanding and commonsense reasoning.
    \item \textbf{XWinograd} (\citep{xwinograd}): A multilingual version of the Winograd schema challenge that tests models on resolving pronoun disambiguation in English, French, and Japanese.
\end{itemize}

\begin{table}[t]
\centering
\small
\setlength{\tabcolsep}{3.3pt} 
\begin{tabular}{l|c|cccccccc}
\toprule
Task &  & \multicolumn{8}{c}{Belebele} \\
Language & Average & \makecell[c]{English \\ (900)} & \makecell[c]{German \\(900)} & \makecell[c]{French \\(900)} & \makecell[c]{Italian \\(900)} & \makecell[c]{Spanish\\ (900)} & \makecell[c]{Chinese \\(900)} & \makecell[c]{Japanese \\(900)} & \makecell[c]{Korean \\(900)} \\
\midrule
Mistral 7B & 0.4073 &  0.4544 & 0.4278 & 0.4222 & 0.3900 & 0.4033 & 0.4067 & 0.3656 & 0.3889 \\
Pythia 7B & 0.3384 & 0.3656 & 0.3267 & 0.3644 & 0.3278 & 0.3300 & 0.3578 & 0.3156 & 0.3189 \\
Wonton 7B & 0.3517 &0.3822 & 0.3500 & 0.3689 & 0.3367 & 0.3467 & 0.3567 & 0.3533 & 0.3189 \\
\bottomrule
\end{tabular}
\caption{Performance comparison of Wonton 7B with other open-source models on the Belebele benchmarks.}\label{tab:result_belebele}
\end{table}

\begin{figure}[t]
\centering
\includegraphics[width=0.6\textwidth]{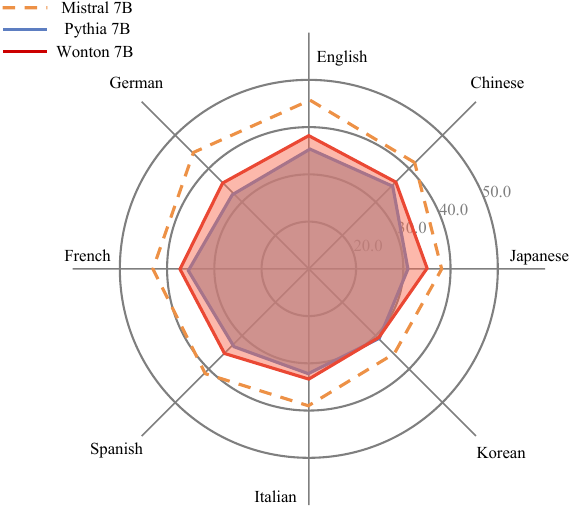} 
\caption{Performance comparison of Wonton 7B with other open-source models on the Belebele benchmarks.} 
\label{fig:belebele}
\end{figure}

\begin{table}[t]
\centering
\small
\setlength{\tabcolsep}{1.5pt} 
\begin{tabular}{l|c|ccc|ccc|ccc}
\toprule
Benchmark & & \multicolumn{3}{c|}{XNLI} & \multicolumn{3}{c|}{XStoryCloze} & \multicolumn{3}{c}{XWinograd}  \\
Language & Average & \makecell[c]{English\\ (5010)} & \makecell[c]{German\\ (5010)} & \makecell[c]{Chinese \\ (5010)} & \makecell[c]{English \\(1511)} & \makecell[c]{Spanish \\ (1511)} & \makecell[c]{Chinese \\ (1511)} & \makecell[c]{English \\ (2325)} & \makecell[c]{French \\(83)} & \makecell[c]{Japanese \\(959)} \\
\midrule
Mistral 7B & 0.6581 & 0.5731 & 0.5026 & 0.3737 & 0.7882 & 0.6903 & 0.6334 & 0.8865 & 0.7470 & 0.7278 \\
Pythia 7B & 0.5841 & 0.5411 & 0.4605 & 0.3387 & 0.7055 & 0.5989 & 0.5381 & 0.8245 & 0.6627 & 0.5871 \\
Wonton 7B & 0.6153 &0.5427 & 0.4884 & 0.3417 & 0.7075 & 0.6453 & 0.6056 & 0.8417 & 0.6627 & 0.7018 \\
\bottomrule
\end{tabular}
\caption{Results of Wonton 7B across XNLI, XStoryCloze, and XWinograd benchmarks.}\label{tab:result_multilingual}
\end{table}

In Table \ref{tab:result_belebele},  Wonton 7B showed competitive results across the Belebele benchmarks. It consistently outperformed Pythia 7B, indicating more efficient use of its training data and better optimization strategies. The results in Table \ref{tab:result_multilingual} are consistent with Belebele, reflecting improvements in its ability to handle complex linguistic constructions and contexts across different languages.

\section{Inference and Deployment}

\subsection{Inference}
In the development of our foundation models, we have implemented and tested several inference frameworks to optimize performance across different environments. Our primary focus was on leveraging the flexibility and robustness of PyTorch for native inference, which allowed us to maintain a consistent development and deployment pipeline.

To expand our model's accessibility and ease of integration, we adapted our models for use with the Hugging Face ecosystem. This adaptation enabled seamless inference capabilities and provided our team with robust, community-driven tools for model management and deployment. The Hugging Face integration also allowed us to leverage their extensive libraries and APIs, which significantly accelerated our development process. This work is yet to be published.

Additionally, we utilized NVIDIA TensorRT\footnote{\url{https://github.com/NVIDIA/TensorRT}} for optimizing our models for high-performance inference on NVIDIA GPUs. This was crucial for scenarios demanding low latency and high throughput, such as real-time applications and large-scale deployments. TensorRT helped in dramatically reducing inference times while preserving the accuracy and reliability of our models.

\subsection{Deployment}
For deployment, we mostly chose Aliyun Elastic Algorithm Service (EAS)\footnote{\url{https://www.alibabacloud.com/help/en/pai/user-guide/eas-model-serving/}} to host and manage our models. This platform offered a scalable and secure environment that supported our operational needs, including automatic scaling, performance monitoring, and robust security measures. The use of Aliyun EAS facilitated a straightforward deployment process and enabled efficient management of model resources in a production environment.

Through these diverse technologies, we achieved a flexible, efficient, and scalable foundation model infrastructure that supports a wide range of applications and meets various business requirements.

\section{Specialized Model for Chat}


\subsection{Data}
We finetune Wonton 7B pretrained model using open source and industry datasets created by Nyonic. All The datasets are in English language. It  includes:
\begin{itemize}
    \item \textbf{Instruction Tuning with GPT-4} (\citep{peng2023instruction}): GPT-4-LLM, which aims to share data generated by GPT-4 for building an instruction-following LLMs with supervised learning and reinforcement learning.
    \item \textbf{Flan V2} (\citep{weifinetuned}): It aims to `Finetuned Language Models are Zero-Shot Learners`.
    \item \textbf{Tulu V2/Cot} (\citep{ivison2023camels}): TÜLU, open resources for instruction tuning have developed quickly, from better base models to new finetuning techniques. TÜLU 2, a suite of improved TÜLU models for advancing the understanding and best practices of adapting pretrained language models to downstream tasks and user preferences.
    \item \textbf{Open Assitant} (\citep{köpf2023openassistant}): It comes in lots of conversations to democratize large language model alignment, especially multi-turn conversations.
    \item \textbf{Automobile Datasets}: It is created by Nyonic and aims to contain knowledge of Automobile industry.
    
\end{itemize}

\subsection{SFT Hypterparameters}
We conduct the full finetuning training based on the Nyonic pretrained model. It leverages the same training framework as the pretraining did. The prompt template is \verb|<s> [INST] Question [/INST] Answer </s>|. We implement the SFT loss function by masking the loss question part, to only accumulate the losses of  the answer part.

The optimizer is similar to the pretraining stage described at \ref{trainparameters}. The initial learning rate is $2\mathrm{e}{-5}$.  We employ a 512 global batch size for smoothing training losses. We conduct 1-3 training epochs.

\subsection{Experiment Results}
\paragraph{Basic Capabilities} 

Table \ref{tab:result_basic} shows the performance of Nyonic pretrained 7B model. The fine tuned model has a comparable results on such benchmarks.

\paragraph{Independent Scoring}

We employs ChatGPT to conduct independent scoring by assigning scores ranging from 1 to10, based on the quality of model responses. The fine tuned model had achieved average 2.6 score higher than pretrained model.

\section{Conclusion}

We have summarized our technical development of our large language models. As we mentioned at the beginning, we will continuously train and release more model checkpoints in the future, as well as fine-tuned models for different scenarios. We hope the community can benefit from our experience and we sincerely welcome further development based on our models and collaborations on various downstream applications.


\medskip


\bibliographystyle{plainnat}
\bibliography{custom}

\end{document}